\documentclass[wcp]{jmlr}

\usepackage{times}
\usepackage{graphicx}
\usepackage{natbib}

\usepackage{longtable}
\usepackage{booktabs}
\jmlrvolume{}
\jmlryear{}
\jmlrworkshop{}

\usepackage[english]{babel}
\usepackage[utf8x]{inputenc}
\usepackage{amsmath}

\usepackage[colorinlistoftodos]{todonotes}
\usepackage{amssymb}
\usepackage{url}

\providecommand{\keywords}[1]{\textbf{\textit{Keywords}} #1}
\providecommand{\e}[1]{\ensuremath{10^{#1}}}
\def\ie{{\it i.e.}}
\def\eg{{\it e.g.}}

\DeclareMathOperator{\ppr}{ppr}

\usepackage{tikz}
\usepackage{pgfplots}
\pgfplotsset{compat=newest}
\def\TT{\mathcal{T}}

 \author{\Name{Jimmy Dubuisson} \Email{jimmy.dubuisson@unige.ch}\\
 \Name{Jean-Pierre Eckmann} \Email{jean-pierre.eckmann@unige.ch}\\
\Name{Andrea Agazzi} \Email{andrea.agazzi@unige.ch}\\
 \addr D\'epartement de Physique Th\'eorique and Section de Math\'ematiques\\ Universit\'e de Gen\`eve }

\title[Diffusion Fingerprints]{Diffusion Fingerprints}

\begin{document}





\maketitle

\begin{abstract}
We introduce, test and discuss a method for classifying and clustering data modeled as directed graphs. The idea is to start diffusion processes from any subset of a data collection, generating corresponding distributions for reaching points in the network. These distributions take the form of high-dimensional numerical vectors and capture essential topological properties of the original dataset. We show how these diffusion vectors can be successfully applied for getting state-of-the-art accuracies in the problem of extracting pathways from metabolic networks. We also provide a guideline to illustrate how to use our method for classification problems, and discuss important details of its implementation. In particular, we present a simple dimensionality reduction technique that lowers the computational cost of classifying diffusion vectors, while leaving the predictive power of the classification process substantially unaltered. Although the method has very few parameters, the results we obtain show its flexibility and power. This should make it helpful in many other contexts.


\end{abstract}


\section{Introduction}

Our method combines several ideas that can be applied in a flexible way, either for classification or graph mining. By using a graph modeling of associations within a data collection, we generate ``fingerprints'' for any subset of data items in the form of high-dimensional distribution vectors of a random walk diffusion process. When such a graph is not provided explicitly, we first  compute association matrices for each subset of the data collection, and then merge them to form a directed graph, after a threshold is applied. Since the diffusion vectors are in a high-dimensional space (the dimension being, for example, the number of different tokens in a corpus of texts), we propose a simple dimensionality reduction that allows for reasonable computational cost associated with the classification of these vectors. 


Various related methods start to build a weighted undirected graph in the form of a similarity matrix, and make use of the spectrum of a derived Laplacian matrix for dimensionality reduction or clustering \cite{shi2000,belkin2001,coifman2005}.

\textit{Diffusion kernels} \cite{kondor2002} propose a general method for constructing kernels on undirected graphs, so that such discrete structures can be used with classical kernel-based learning algorithms \cite{scholkopf2002}. Unfortunately, this approach remains in general computationally expensive for large graphs. 

Random walks over undirected graphs have also been used as a similarity measure for collaborative recommendation \cite{fouss2007}, community detection \cite{andersen2006,pons2006}, or relational classification \cite{perozzi2014}. The \textit{ItemRank} algorithm \cite{gori2007}, on the other hand, uses biased random walks over a directed graph to compute a score vector for each user, within the framework of a recommender system. 

Although related to these previous works, our method combines a set of specific features. First, by operating preferably on directed graphs, it enables to get deeper insights into the topology of the dataset under consideration. Moreover, it can potentially use many types of biased random walks, and does not need the diffusion processes to reach full convergence (this may help to quickly compute estimates of the fingerprints, or to get ``snapshots'' of the dataset at different times). Finally, it can be used to compute numerical vectors (one may think of ``feature vectors'') for any subset of data items, including overlapping ones for example.

In the present paper, we start by presenting the basic formalism of our method in section~\ref{formalism}, then show how it can be successfully applied to the problem of extracting relevant pathways from metabolic networks in section~\ref{mpi}. In section~\ref{nlp}, we provide a guideline to illustrate how to use our method for classification problems. Finally, we discuss a few important details more thoroughly in section~\ref{discussion}.

\section{Formalism}
\label{formalism}

\subsection{The association matrices}

We consider a \textit{data collection} 
$\Sigma=\{\sigma(1),\sigma(2),\dots\}$ of ``documents'', where each \textit{document} is viewed as a set of data items whose pairs get assigned a value of association. Some data collections such as the USF Free Associations dataset \cite{fa98,dubuisson2013} already provide explicit association weights while for others, one needs to define a way to compute these association values. The set of different items appearing in $\Sigma$ is called $\TT$. We denote $|\TT|$ the cardinality of $\TT$. 

\subsubsection{An example: computing words collocation}

To illustrate our terminology and give an example of how to compute such association values when necessary, we consider a corpus $\Sigma$ of text documents. A document $\sigma(k)$ will consist here of $N(k)$ tokens, which are typically stemmed words, with some stop words omitted. We define $I(k)=\{I_k(1),\dots,I_k(N(k))\}$ as the list of tokens of $\sigma(k)$  in the order in which they appear. Each $I(k)$ is thus a map from positions in $\sigma(k)$ to tokens in $\TT$. 

We next define, for each $k$, the association matrix $K(k)$, which is
a $|\TT|\times|\TT|$ matrix. We fix $k$ and omit the index $k$ for the
moment. The matrix $K$ measures the association of pairs of tokens
$u,v\in\TT$. For every $u\in\TT$, we let $p_u(i)$ be the position of
the $i^{\rm th}$ occurrence of token $u$. For every ordered pair
$(u,v)$ with $u\ne v$ of tokens we look for occurrences of the form  
$p_u(i) < p_v(j)< p_u(i+1) $, that is, occurrences of token $v$ between two successive occurrences of token $u$ (or after the last occurrence of $u$). We let $s_{uv}$ be the set of all such pairs $(i,j)$. Still
omitting the index $k$, the matrix $K$ is defined by 

$$
K_{uv} = g(h(u,v)) \sum_{(i,j)\in s_{uv}} f(p_u(i),p_v(j))~,
$$
where
$$ h(u,v) = \frac{|s_{uv}|}{\sum\limits_{\substack{u',v' \in
\TT(k) \\ u' \neq v'} } |s_{u'v'}|}~,
$$
$\TT(k)$ are the tokens appearing in $I(k)$ and the functions $g(\cdot)$ and $f(\cdot,\cdot)$ are defined below. Note that $K_{uv}=0$ if
$s_{uv}$ is the empty set.

We conduct our experiments with the following families of functions:
$$ f(i,j) = \exp(-\frac{(j-i-1)^\beta}{\sigma}) \text{ and }
 g(x) = -\log(x)~. $$
The rationale behind the use of the function $f$ is that the
collocation measure should decrease exponentially with the distance
between any two tokens. $g(h(\cdot))$ is a function of the relative
frequency of each pair $s_{uv}$ and serves as a normalizing function
whose goal is to correct the influence of very frequent pairs. (These
functions might be changed somewhat depending on the study one wants
to perform.)

\subsection{The domain graph}

Having generated, for each document $k$, the association matrix $K(k)$
as described above, we next define a matrix
$K(\Sigma)$ for the whole data collection $\Sigma$ by $$
K(\Sigma)_{uv} = {\sum_k} K(k)_{uv}~.$$

We now introduce a density parameter $\gamma$ and define with it an
\textit{adjacency matrix} $A(\gamma)$ as follows: we replace
the $N\equiv\gamma|\TT|\cdot(|\TT|-1)$ largest elements of $K(\Sigma)$
by 1, and the others by 0. This means that the matrix elements of
$K(\Sigma)$ above a certain threshold are replaced by 1 and the others
by 0.\footnote{In case of multiplicities (for example if all matrix elements of $K(\Sigma)$ are equal), we perform a random choice of the required number of elements.} Note that our method does not specifically require the use of a binary adjacency matrix, and can easily be adapted to work with a weighted one, if it brings a clear benefit to do so.

The \textit{domain graph}
$G(\gamma)$ is the directed graph whose nodes are the elements of
$\TT$ and whose adjacency matrix is
$A(\gamma)$.  The topology of $G(\gamma)$ reflects the $N$
strongest associations in $\Sigma$ 
for a given density $\gamma$.

\subsection{The diffusion fingerprints}

Having determined the directed graph $G(\gamma)$, we now consider a diffusion process on it. In particular, we are interested in how a given document $\sigma(k)$ fits into this graph.

For a fixed $k$, there is a set $\TT'(k)\subset\TT$ of data items
which are nodes of the domain graph and which appear in $\sigma(k)$.
We want to know how the set $\TT'(k)$ diffuses into the domain
graph.\footnote{$\TT'(k)$ might be smaller than the set $\TT(k)$ of
tokens in $\sigma(k)$.}  

We call \textit{diffusion fingerprint} (DF) of document $\sigma(k)$ the distribution vector of the diffusion process started from the subset $\TT'(k)$ of nodes
in $G(\gamma)$. Note that the smaller the set $\TT(k)\backslash\TT'(k)$, the better the generated fingerprint represents $\sigma(k)$ within the context of the domain graph. 

Let $P$ be the \textit{probability matrix} defined by

$$ P(\gamma) = D^{-1}(\gamma)A(\gamma)~, $$
where $D$ is the diagonal matrix of the degrees of $G$.
We compute the DF of document $\sigma(k)$ as the
\textit{personalized Pagerank} \cite{page1999,Andersen:2006:LGP:1170136.1170528} $\ppr_k$ defined
recursively by
\begin{equation}
\ppr_k(t+1) = \alpha v_k + (1-\alpha)\ppr_k(t) P~,
\end{equation}
where $\alpha \in (0,1]$ is called the \textit{jumping constant} and
$\ppr_k(0)=v_k$. 
The vector $v_k$ is the \textit{personalized vector} given by

$$ 
v_k(u) = \left\{
    \begin{array}{ll}
       f_k (u) & \mbox{if } u \in \TT(k) \\
        0 & \mbox{otherwise}
    \end{array}
\right. ,
$$

with $f_k(u)$ the frequency of data item $u$ in document $\sigma(k)$.
Note that the parameter $(1-\alpha)$ is the inverse of the expected path length of a random walker before being projected back to $\TT'(k)$.\\
In principle, we define

$$ \pi(k) = \lim_{t \to \infty} \ppr_k(t)~,$$
and call $\pi_t(k)$ the \textit{DF at time t} of document $\sigma(k)$, and $\pi(k)$  its \textit{stationary DF}.\footnote{Of course, we just compute $\ppr_k(t)$ for some sufficiently large $t$.} Note that by default, we use  the personalized Pagerank algorithm for computing the diffusion vectors as its properties are well-studied and understood, but nothing prevents our method to be used with other types of biased random walks. Considering the DFs at different times may also permit to construct derived feature vectors taking into account the dynamics of the diffusion processes.

\section{Application to metabolic pathway inference}
\label{mpi}

We will now apply the general method sketched out in section~\ref{formalism} to the 
problem of extracting metabolic pathways from metabolic networks. After describing what the question is, we proceed with the application of the general ideas, adapting them, where needed, to the specificities of the task.

\subsection{Description of the problem}

The understanding of the dynamics underlying the set of metabolic reactions in a living cell has been pushed a step further by the extensive application of high-throughput analysis techniques in cell biology. In particular, the quantity and accuracy of data produced with these methods has dramatically increased in the last decade, and new computational techniques are required in order to interpret the set of these experimental results as a whole. 

The metabolomic datasets can be represented by simple or bipartite graphs. In the first case, the nodes of the graphs are the metabolites, and a (directed or undirected) link is drawn between two molecules if there exists a chemical reaction having one of them as substrate and the other as product. In this case, we refer to the graph as the species-species graph (SSG). In the second case, the nodes of the graphs representing chemical interactions are both chemical molecules and reactions, interlinked by in- and out-flow relations. This kind of representation is referred to as species-reactions graph (SRG). An interesting problem in this context is the one of predicting metabolic pathways (chains of interlinked reactions, transforming a set $S$ of source metabolites into a set $T$ of targets) in metabolic databases.

Different approaches have been proposed for solving this problem: \cite{Zien00} have developed a method that ranks paths connecting a source and a target node in a SRG, according to gene expression levels corresponding to enzymes catalyzing reactions on the possible paths. Approaches based on the same intuition have been developed for pathway discovery in protein-protein interaction networks, by applying breadth-first search-based and Steiner tree problem solving algorithms \cite{Scott05,Rajagopalan05}.

Other algorithms for metabolic pathway extraction based on shortest-path finding have been developed in \cite{Antonov09,Faust10}. In particular, a method based on random walks on weighted metabolic graph is used in \cite{Faust10} to extract metabolic pathways connecting a given set of nodes in a SRG extracted from the \textit{MetaCyc v11.0} database. This is done by searching for the shortest paths interlinking the set of terminal nodes of a given pathway, where by shortest path one means the path with minimal summed weights. 

The weighting algorithm used in this case is based on the expected number of times a random walker transits through a given edge before reaching a terminal node of the pathway. The computational cost of this algorithm is $O(sm^3)$ \cite{Dupont06}, whereby $m$ is the total number of edges of the domain graph and $s$ the cardinality of $S\cup T$). An arbitrary limit on the path length must be given in order to reduce the complexity of the algorithm. In the following we propose an alternative to this metabolic pathway-finding algorithm, based on the DF method explained above.

\subsection{Description of the algorithm}

The graph $G$ here is the directed domain graph modeling the metabolic network we want to analyze and $P$ is the corresponding transition probability matrix. No filtering procedure has been applied on the domain graph in this case. We call \textit{reverse graph} of $G$ and denote $G^*$ the graph having the same set of vertices as $G$, with all its edges reversed. We thus have $P^* = P^t$.

The domain graph $G$ can be either simple or bipartite according to the representation of the metabolic network we choose to apply (SSG or SRG respectively). 

Let $R \subset V$ be the set of species that participate in a given annotated metabolic pathway. The nodes in $R$ are weakly connected in the domain graph $G$. We also denote $S \subset R$ the set of sources (\ie, nodes with null in-degree), and $T \subset R$ the set of sinks (\ie, nodes with null out-degree) of the pathway.

For a general pathway, we want to reconstruct the set $R$ of nodes participating in it starting from the sets $S$ of sources and $T$ of sinks. 

We start by computing $\pi(S)$, the stationary DF of $S$ in $G$, and $\pi^*(T)$, the stationary DF of $T$ in $G^*$. In order to exhibit the set of nodes that are highlighted by both of the fingerprints, we consider the \textit{combined diffusion fingerprint}
$$\pi_{\triangleright}(S,T) = \pi(S) \times \pi^*(T)~,$$ 
where $\times$ represents the Hadamard Product (component-wise multiplication).





\subsection{Pagerank boosting}

Because of the presence of hub nodes (such as $H_2O$, \textit{ADP}, \textit{NADH}) in the corresponding graphs, a direct application of the algorithm described above would result in pathways connecting source and target nodes only through such highly connected compounds, since they effectively represent the shortest path to connect any given set of two nodes in the graph. This problem is well known and different methods have been developed to overcome it \cite{Faust11}. We propose here a method which is softer than the direct elimination of hub metabolites having a total degree above an arbitrarily fixed threshold.

In order for the algorithm to find relevant metabolites, \ie, those that characterize a particular pathway and belong to as few other paths as possible, we renormalize our results keeping the centrality of the different nodes into account. In particular, we rescale the values of the combined diffusion fingerprint vector $\pi_{\triangleright}(S,T)$ resulting from the algorithm described above using the Pagerank vector $\pi(G)$ of the full graph $G$, and the Pagerank vector $\pi(G^*)$ of the full reverse graph $G^*$ in the following way:
$$\pi_{\triangleright}(S,T)^{boosted} = \frac{\pi_{\triangleright}(S,T)}{\pi(G) \times \pi(G^*)~},$$
where both multiplication and division are intended in the component-wise sense.

This boosting procedure effectively penalizes metabolic ``hubs'' such as $H_2O$ and \textit{ATP} which have high in- and out-degrees.

\subsection{Pathway selection}

In order for the algorithm to detect a pathway connecting a set $S$ of source nodes to a set $T$ of sink nodes, we consider the $n$ largest entries of the vector $\pi_{\triangleright}(S,T)^{boosted}$, and increase $n$ until the subgraph resulting from the first $n^{w}$ compounds connects all the elements of $S$ to all the elements of $T$ in the weak sense. The extracted subgraph (the inferred pathway) is then given by the subset of the $n^w$ largest entries of $\pi_{\triangleright}(S,T)^{boosted}$ that belong to the weakly connected component connecting $S$ to $T$.

\subsection{Application of the algorithm}

We applied the algorithm described above to the problem of finding the set of known pathways in the domain graph extracted from the annotated chemical reactions in the \textit{MetaCyc v18.5} database. The domain graph is an SSG of $9\,553$ nodes and $75\,078$ edges. In order to have realistic results, we have applied the algorithm to the search of pathways of length $l \geq 3$, where $l$ is the minimal shortest path length between any source and target node in the annotated pathway. In total, the algorithm has been applied to the reconstruction of $1\,981$ pathways.

%
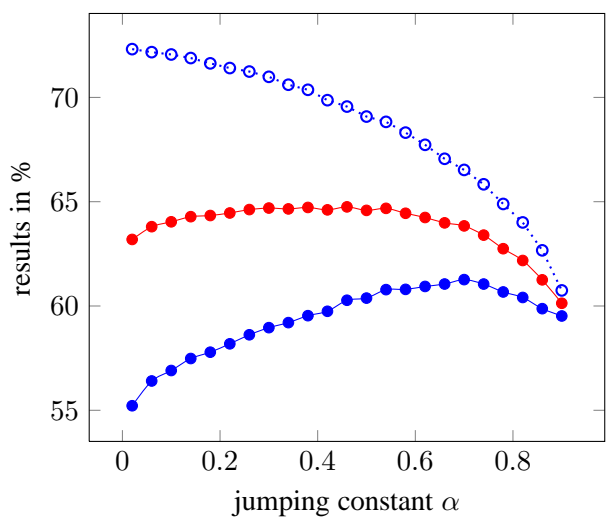
\begin{figure}[t]
	\centering
	\begin{tikzpicture}[scale=1]
    \begin{axis}[
        xlabel=jumping constant $\alpha$,
        ylabel=results in \%]
    \addplot[smooth,mark=*,red] plot coordinates {
        (0.02,63.1905195895)
        (0.06,63.8024599203)
        (0.10,64.0359056927)
        (0.14,64.2821830775)
        (0.18,64.3373833602)
        (0.22,64.4588788822)
        (0.26,64.6225534424)
        (0.30,64.6955787829)
        (0.34,64.6544245841)
        (0.38,64.7245307309)
        (0.42,64.6089192264)
        (0.46,64.7522618967)
        (0.50,64.5837580685)
        (0.54,64.6812747491)
        (0.58,64.4451047738)
        (0.62,64.2426721092)
        (0.66,63.9852170133)
        (0.70,63.8401701376)
        (0.74,63.398004195)
        (0.78,62.7467936616)
        (0.82,62.1813519684)
        (0.86,61.2537276525)
        (0.90,60.1308454322)
      };
      \addplot[smooth,mark=*,blue] plot coordinates {
        (0.02,55.2175289762)
        (0.06,56.4051962384)
        (0.10,56.9071317893)
        (0.14,57.4818590854)
        (0.18,57.7853285749)
        (0.22,58.1863491712)
        (0.26,58.6198650879)
        (0.30,58.9617487584)
        (0.34,59.2026696499)
        (0.38,59.5360702613)
        (0.42,59.7460368161)
        (0.46,60.2757380293)
        (0.50,60.3784100192)
        (0.54,60.784653951)
        (0.58,60.7971607377)
        (0.62,60.937262368)
        (0.66,61.0533116646)
        (0.70,61.2648787401)
        (0.74,61.0519440588)
        (0.78,60.6726072817)
        (0.82,60.4086598729)
        (0.86,59.8718184643)
        (0.90,59.5210639613)
      };
      \addplot[dotted,thick,mark=o,mark options={solid},blue] plot coordinates {
        (0.02,72.31474932)
        (0.06,72.1698383014)
        (0.10,72.0577032957)
        (0.14,71.887011432)
        (0.18,71.632350282)
        (0.22,71.4075917451)
        (0.26,71.2399185352)
        (0.30,70.9870043239)
        (0.34,70.6082114711)
        (0.38,70.3651561139)
        (0.42,69.8676040464)
        (0.46,69.5612456657)
        (0.50,69.0820080377)
        (0.54,68.8276897412)
        (0.58,68.3119323157)
        (0.62,67.7273766387)
        (0.66,67.057918475)
        (0.70,66.523714843)
        (0.74,65.8342170404)
        (0.78,64.8918892925)
        (0.82,64.0060637127)
        (0.86,62.6675328654)
        (0.90,60.7468739931)
      };
         \end{axis}
    \end{tikzpicture}
    \caption{Precision (filled blue circles) and recall (empty blue circles) of pathway inference averaged over $1981$ experiments as a function of the jumping constant $\alpha$. The geometric mean (red filled circles) of these two quantities, later also referred to as the {\it base score geometric accuracy}, is very weakly affected by the choice of $\alpha$ in the interval $\alpha \in (0.1,0.6)$.
}
	\label{fig:PIAccuracy}
\end{figure}


We have tuned the only parameter of our algorithm to the value $\alpha = 0.15$, as generally assumed in the Pagerank literature \cite{Brin98}. However, our results are surprisingly invariant with respect to variations of this parameter. This behavior can be explained as follows. For a jumping constant $\alpha \approx 1$, the expected path length approaches the value $0$ and no pathway can be inferred. On the opposite case, for $\alpha \approx 0$, one approaches the limit of a standard random walk, and all pathways, independently on their length, will be indistinguishably found. However, when we are in neither of these two cases, since the node weights in the $\pi_\triangleright(S,T)$ vector exponentially decrease as a function of distance from the source and target sets, shorter pathways will \textit{always} be preferred by our algorithm over longer ones. After that, applying pagerank boosting, hub nodes are penalized and only the relevant shortest paths are evidenced.
The behavior explained above is displayed in Fig.~\ref{fig:PIAccuracy}. There we see that the geometric mean of precision (\textit{PPV}) and recall (\textit{TPR}) of our results is very weakly influenced by the choice of the parameter $\alpha$, the latter two measures being defined as
$$
\textit{PPV} = \frac{\textit{TP}}{\textit{TP} + \textit{FP}}, \quad\textit{TPR} = \frac{\textit{TP}}{\textit{TP} + \textit{FN}}
$$
where \textit{TP} is the number of non-seed nodes that are present both in the inferred and in the annotated pathway. \textit{FP} is the number of nodes that are present in the inferred but not in the annotated pathway, and \textit{FN} is the number of nodes present in the annotated but not in the inferred pathway. This weak dependency makes the highly parallelizable algorithm presented above virtually parameter-free.

\subsection{Results and discussion}


To our knowledge, the strategy appearing in the bioinformatics literature that relates most closely to ours is the one presented in \cite{Faust10}, where a random walk approach is used to infer metabolic pathways in metabolomics databases. There, the parameter chosen to quantify the quality of the results is the \textit{geometric accuracy}, defined as
$$\textit{acc}_g = \sqrt{\textit{PPV}\cdot \textit{TPR}}~.$$ In the following we will use the same parameter in order to simplify the comparison of the results, which is nonetheless difficult for several reasons.

The first is that even though the metabolomic database on which the algorithm has been applied is the same, \textit{MetaCyc} has been updated multiple times since $2010$, doubling in size. Therefore, the graph on which we have applied our algorithm is different from the one used in the article to which we are comparing.

The second reason is that the research of pathways in the article mentioned above is carried out by diffusion starting from source subgraphs $S$ of increasing cardinality until the whole annotated pathway is covered. The result for the geometric accuracy is then given by the average of the geometric accuracies for each diffusion. In our case the set $S$ is given by the terminal nodes of the pathway only.

Noting these differences, we have compared our results to those obtained in \cite{Faust10} selecting out of all the pathways analyzed therein those that are still present in the latest version of the \textit{Metacyc} database. The comparison has been carried out by applying our DF-based algorithm to the inference of all pathways contained in the \textit{Metacyc v18.5} database.

We can reconstruct the value of the geometric accuracy for the algorithm we are comparing to in the case only terminal nodes are taken as seed nodes. 
We define this parameter as the $\textit{base score geometric accuracy: acc}_g^b$. In this case, for the pathways that are still present in the updated version of the database, we find a value of $\textit{acc}_g^b = 0.61 \pm 0.05$. The uncertainty of $\pm 5 \%$ is due to difficult decoding of the published data. The geometric accuracy of the inference of metabolic pathways using the DF-based algorithm described above is $\textit{acc}_g^b = 0.66$.

\section{A guideline to use DF for classification}
\label{nlp}

This section is dedicated to further illustrating our method within the context of two classical textual classification problems, and provide a guideline to apply DF to classification tasks.

\subsection{Gender detection}

We first apply our method to a binary textual classification problem. Given a set of about $20\,000$ blogs \cite{schler2006}, our goal is to determine the gender of the authors \cite{df14}. 

We randomly select a subset of $1\,000$ blogs with an equal proportion of male and female authors and use it as a training set. In both examples, we follow the procedure described in section~\ref{formalism} for generating the association matrices. Setting the density parameter $\gamma=\e{-2}$ and the function $f$ parameters $\beta = \sigma = 1$, we compute the domain graph $G$ and the set of fingerprints for this subset.
The graph $G$ has $23\,629$ nodes and $5\,583\,061$ edges. At this specific density, the graph forms a single strongly connected component and has a directed diameter equal to $6$.  

\begin{figure}[t]
	\centering
	\begin{tikzpicture}[scale=1]
    \begin{axis}[
        xlabel=density $\gamma$,
        ylabel=accuracy in \%]
    \addplot[smooth,mark=*,blue] plot coordinates {
        (0.0001,73.91)
        (0.001,76.87)
        (0.002,76.25)
     	(0.003,75.39)
        (0.004,75.81)
        (0.005,77.08)
        (0.006,76.24)
        (0.007,75.99)
        (0.008,77.01)
        (0.009,77.04)
        (0.01,78.07)
        (0.011,77.29)
        (0.012,77.02)
      };
      \addplot [red, no markers, densely dotted] coordinates {(0.003,74) (0.003,78)};
      \addplot [red, no markers, densely dotted] coordinates {(0.011,74) (0.011,78)};
    \end{axis}
    \end{tikzpicture}
    \caption{Accuracy of gender detection as a function of the density parameter
      $\gamma$. The left red vertical line marks the minimum density
      above which the domain graph becomes a strongly connected
      component (\ie, the diffusion process can reach all parts of the
      graph). The right red vertical line marks a density above
      which the accuracy starts decreasing and the computation of
      fingerprints becomes very costly.}  
	\label{fig:binaryAccuracy1}
\end{figure}
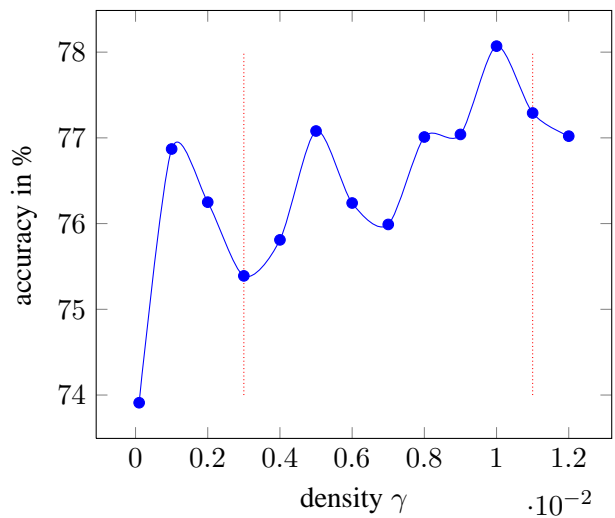

We then perform 10-fold cross validation on $10$ random shufflings of
the $1000$ fingerprints. We get an average accuracy of $79.1\%$ with
the AdaBoost meta-algorithm using Decision Tree classifiers
\cite{Freund:1995:DGO:646943.712093}. By comparison, we get an average
accuracy of $74.8\%$ when we use simple bag-of-words (BOW) vectors on the
same set of blogs. Note that we use a BOW model not as a model to compete with, but rather as a null model to compare the effects of our dimensionality reduction heuristic (see details in section~\ref{discussion}). 

Moreover, when we apply our \textit{OPC} dimensionality reduction heuristic to the fingerprint vectors, we observe that the accuracy remains almost constant until we reach a dimension equivalent to $10\%$ of the size of the domain graph. For instance, the accuracy still only reduces slightly to $77.85\%$ when the dimension $d$ is reduced from $23\,629$  to $3\,000$. 


%
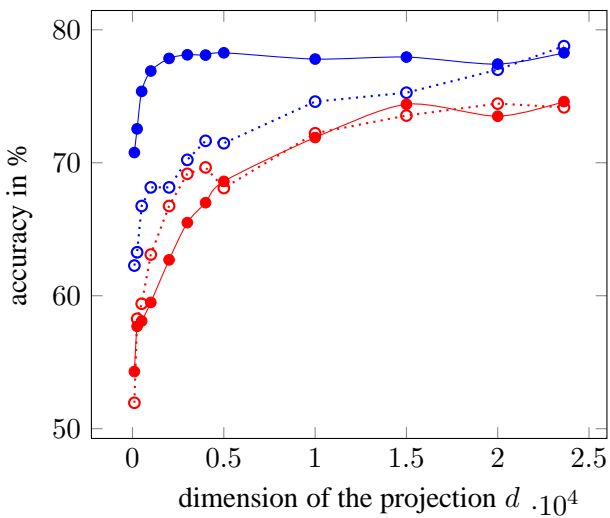
\begin{figure}[ht]
	\centering
	\begin{tikzpicture}[scale=1]
    \begin{axis}[
        xlabel=dimension of the projection $d$,
        ylabel=accuracy in \%]
		%
    \addplot[smooth,mark=*,blue] plot coordinates {
        (100,70.77)
        (250,72.55)
        (500,75.37)
        (1000,76.9)
        (2000,77.85)
        (3000,78.12)
        (4000,78.1)
        (5000,78.27)
        (10000,77.8)
        (15000,77.95)
        (20000,77.42)
        (23629,78.27)
      };
      \addplot[dotted,thick,mark=o,mark options={solid},blue] plot coordinates {
        (100,62.27)
        (250,63.27)
        (500,66.75)
        (1000,68.15)
        (2000,68.15)
        (3000,70.22)
        (4000,71.65)
        (5000,71.47)
        (10000,74.6)
        (15000,75.27)
        (20000,77)
        (23629,78.77)
      };
      \addplot[smooth,mark=*,red] plot coordinates {
        (100,54.3)
        (250,57.7)
        (500,58.1)
        (1000,59.5)
        (2000,62.7)
        (3000,65.5)
        (4000,67)
        (5000,68.6)
        (10000,71.9)
        (15000,74.4)
        (20000,73.5)
        (23629,74.6)
      };
      \addplot[dotted,thick,mark=o,mark options={solid},red] plot coordinates {
        (100,51.95)
        (250,58.27)
        (500,59.4)
        (1000,63.1)
        (2000,66.75)
        (3000,69.17)
        (4000,69.65)
        (5000,68.1)
        (10000,72.22)
        (15000,73.55)
        (20000,74.45)
        (23629,74.17)
      };
    \end{axis}
    \end{tikzpicture}
    \caption{Accuracy of gender detection as a function of the reduced
      dimension $d$ for $\gamma = \e{-2}$. The curves are: the diffusive fingerprint method with OPC (solid blue) and the diffusive fingerprint method with random projection (dotted blue), bag of words with OPC (solid red), and bag of words with random projection (dotted red). Note the stability of the fingerprint-OPC result when the dimension $d$ is lowered.
}
	\label{fig:binaryAccuracy2}
\end{figure}

\subsection{Authorship attribution}

By using the same set of blogs, we also apply our method to the problem of authorship attribution \cite{ko11,Seroussi:2012:AAA:2390665.2390728}. We start by selecting at random $500$ blogs containing at least $16$ posts of more than $8$ tokens each, and we split each of them in two equal number of posts. 

For the $500$ selected blogs (\ie, 500 classes), we get $11\,993$ posts containing more than $8$ tokens. We first use the aggregated list of tokens of the first halves for generating the domain graph. By choosing $\gamma=\e{-2}$ and setting $\beta = \sigma = 1$, we get a graph with $17\,036$ nodes and $2\,902\,681$ edges. We then generate the DFs for each post, and use the fingerprint vectors of the first halves for training $500$ `one-vs-all` Random Forest binary classifiers \cite{RF}.

We finally use the fingerprint vectors of the posts in the second parts for testing our classifier, which gives us an accuracy of $27.6\%$. By comparison, the accuracy is reduced to $24.2\%$ when we use simple BOW vectors.

It is to be noted that as the number of classes increases, the computation needed to train the binary classifiers can become very costly. One way to alleviate this problem is to compute instead the mean of the training fingerprint vectors, and use it as a reference fingerprint for each author (\ie, class). When the Manhattan distance is used for classifying the test vectors, we get an accuracy of $22.25\%$ for the DFs and $9.36\%$ for the corresponding BOW vectors.

%
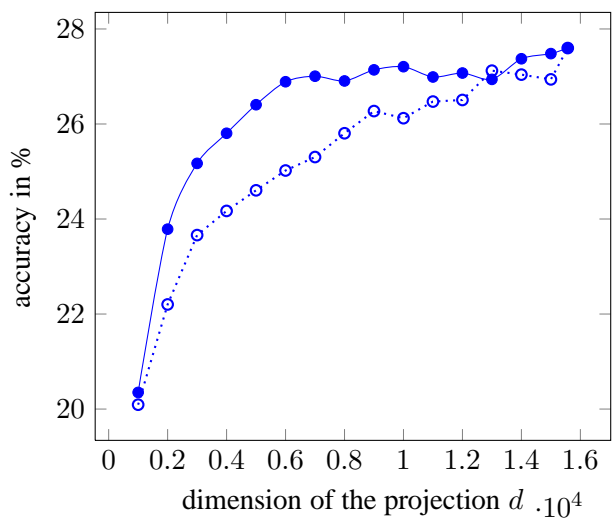
\begin{figure}[ht]
	\centering
	\begin{tikzpicture}[scale=1]
    \begin{axis}[
        xlabel=dimension of the projection $d$,
        ylabel=accuracy in \%]
    \addplot[smooth,mark=*,blue] plot coordinates {
        (1000,20.35029191)
     	(2000,23.78648874)
        (3000,25.17097581)
        (4000,25.80483736)
        (5000,26.40533778)
        (6000,26.88907423)
        (7000,27.0058382)
        (8000,26.9057548)
        (9000,27.13928274)
        (10000,27.206005)
        (11000,26.98915763)
        (12000,27.07256047)
        (13000,26.93911593)
        (14000,27.37281068)
        (15000,27.48123436)
        (15572,27.59799833)
    };
    \addplot[dotted,thick,mark=o,mark options={solid},blue] plot coordinates {
        (1000,20.09174312)
        (2000,22.20183486)
        (3000,23.66138449)
        (4000,24.17014178)
        (5000,24.60383653)
        (6000,25.02085071)
        (7000,25.30442035)
        (8000,25.80483736)
        (9000,26.27189324)
        (10000,26.12176814)
        (11000,26.47206005)
        (12000,26.50542118)
        (13000,27.12260217)
        (14000,27.03919933)
        (15000,26.93911593)
        (15572,27.59799833)
      };
    \end{axis}
    \end{tikzpicture}
    \caption{Accuracy of authorship attribution as a function of the reduced dimension $d$. The curves are: the diffusive fingerprint method with OPC (solid blue) and the diffusive fingerprint method with random projection (dotted blue). Note that BOW results are not indicated here for sake a clarity.} 
	\label{fig:multiAccuracy}
\end{figure}

\subsection{Results and discussion}

We emphasize that we use text classification problems as a mere illustration of our method. While this does not lead to state-of-the-art accuracies, it shows that the diffusion fingerprints method is generic enough to achieve quite decent results.

\subsubsection{Gender detection}

In \cite{schler2006}, the authors use a dataset of $37\,478$ blogs with an equal number of male and female bloggers to perform gender classification. When applying 10-fold cross validation, they report an accuracy of 72\% with content based features only, and 80.1\% when stylistic features are also included.

By using many more features (\ie, F-measure, stylistic features, gender preferential features, factor analysis, word classes and POS sequence patterns) and a custom \textit{Ensemble Feature Selection} (EFS) algorithm, \cite{mukherjee2010} report a maximum accuracy of 88.56\% over a set of $3\,100$ blogs.

Thus, the accuracy of 79.1\% we report seems acceptable, considering that we use no specific tools like n-grams and no additional features.

\subsubsection{Authorship attribution}

In \cite{ko11}, a dataset of $10\,000$ blogs is used for testing a method of authorship attribution over a large set of classes. For their first experiment, each blog is split in a reference text of $2\,000$ words and a snippet of 500 words, both represented as vectors of space-free character 4-gram frequencies. Using cosine similarity, 46\% of $1\,000$ snippets selected randomly get assigned correctly. They also propose an improved algorithm consisting in repeatedly selecting a random fraction of the full feature set, in a similar way to classifier ensemble methods \cite{bryll2003}.

In our experiment, we obtain a lower accuracy of 27.6\%. However, it is to be noted that in our case, the blog posts to be classified can contain as few as $8$ tokens, which makes the classification task somewhat harder.

\section{General discussion}
\label{discussion}


By contrast to some non-linear dimensionality reduction techniques
which aim at discovering the structure of a manifold from a set of
feature vectors \cite{Roweis00nonlineardimensionality,Tenenbaum2000,Belkin03laplacianeigenmaps,Nadler05diffusionmaps}, DFs start by extracting the latent topological properties of the provided data under the form of a directed
graph. The distributions of the diffusion processes started from the data
subsets are then computed as numerical vectors which can subsequently be fed to any classification algorithm.

\subsection{Choosing the density parameter $\gamma$}

We discuss here how to  determine the density that provides
the best accuracy.

We note that when $\gamma=0$, there is no
diffusion and our method amounts to classifying the  personalized
vectors associated to each subset in a high dimensional space. In this
case, our method simply corresponds to using a bag-of-words model. 

On the other hand, when $\gamma=1$ (then the graph is
complete), the initial distributions (\ie, personalized vectors) get
flattened, whereas all the other coordinates of the high dimensional
fingerprint vectors get assigned the same value. Thus, in this case,
the fingerprint stationary distributions resemble an attenuated
form of the initial distributions (\ie, the personalized vectors), except that the variance of the data is much lower. 

Diffusing in the domain graph from the initial distributions enables
to grasp the generic characteristics of the data subsets at the scale
of the whole domain graph, and improves average accuracy. A first
requirement is thus that the density of the domain graph is large
enough (\ie, greater than a critical value $\gamma_c$) to enable a
diffusion process starting from any subset of nodes to reach all the
other parts of the domain graph. This means that $\gamma$ needs to be
chosen to enable the emergence of a giant strongly connected component
(SCC). We see for example that the domain graph used previously for
gender detection forms a single SCC for $\gamma \geq 3\cdot\e{-3}$. 

We observe that the average entropy of the fingerprint distributions increases monotonically with the graph density, whereas, at the same time, the average variance decreases. Moreover, the average accuracy appears to be approximately a concave function of the domain graph density. In order to reach the best possible accuracy, we thus need to choose the density parameter $\gamma \in [\gamma_c, 1)$ so that the
generated fingerprint distributions retain a high variance, but also exhibit a high average entropy: this corresponds to seeking a trade-off
between the expressiveness and the genericity of the generated vectors. 

\subsection{Dimensionality reduction}

After the construction of the diffusion vector, we use the following procedure for dimensionality reduction to lower the computational cost of classifying the generated DF.  

Fix some $d$ and let $\Phi(d): \mathbb{R}^{|\TT|} \to \mathbb{R}^d$
be the orthogonal projection of the $|\TT|$-dimensional fingerprint
vectors onto the $d$-dimensional node subspace $\mathbb{R}^d$ ($d \ll |\TT|$) spanned by the $d$ most central nodes of the domain graph $G\equiv G(\gamma)$. 
To find this projection, we apply the \textit{Pagerank} centrality metric \cite{Brin98} to $G$ in order to determine the set of the $d$ most central nodes, corresponding a set $\TT_d$ of tokens.
The projection $\Phi(d)$ is then the $d \times |\TT|$-matrix defined by
$$ \Phi(d)_{uv} = \left\{
    \begin{array}{ll}
        1 & \mbox{if } v\in\TT_{d} \\
        0 & \mbox{otherwise}
    \end{array}
\right. ~,\quad u\in\TT_d, v\in\TT~.$$
The intuition is that projecting the fingerprint vectors onto the $d$
most central nodes amounts to embedding the data in a $d$-dimensional
hyperplane of maximum variance, as we will see next.

We call \textit{OPC} (Orthogonal Projection on Central nodes) the
projection we just described.

\subsection{OPC dimensionality reduction heuristic}


Applying dimensionality reduction by projecting orthogonally on the hyperplane spanned by the set of most central nodes limits the decrease of classification accuracy and is computationally very efficient. 

By using the Pagerank metric as a measure of centrality, this
heuristic works well because the highest components of the Pagerank
vector are highly correlated with the fingerprint coordinates of
maximum variance. In the case of the domain graph applied above
for gender detection, the Spearman correlation between the Pagerank
distribution and the vector of fingerprint variance per coordinate is
for example equal to $0.92$ for $\gamma = \e{-2}$. This means that
projecting on the most central nodes amounts to projecting
orthogonally on the hyperplane of maximum variance \cite{PCA,Hotel}. 

The diffusion process plays a fundamental role in allowing a  graceful
decay of the classification accuracy when the dimension of the
fingerprint vectors is reduced by applying the heuristic we just
described. By extracting the generic characteristics of the data
subsets at the scale of the whole data collection, diffusion
fingerprints are in this case far more resilient to dimensionality
reduction compared to a bag-of-words model. 

\subsection{Computational considerations}

The first step of our method consists in generating the association
matrices for the set of documents $\sigma(k)$. Here, the amount of
computation highly depends on whether the association weights are provided in the data (\eg, USF Free Associations dataset \cite{fa98}) or not.

The generation of the domain graph is then straightforward, but we may
face a problem if the number of data items is too large for the domain
matrix to fit into memory. One potential way to alleviate the problem
and to avoid explicitly computing the domain matrix is to generate for
each association matrix a labeled unweighted directed graph of density
$\gamma$ in the form of a sparse binary adjacency matrix, and to
generate the domain graph by taking the edge-union of the set of
labeled subgraphs (\ie, $G = \bigcup E(G_k)$). Note that in this case, it
is more difficult to control the overall density $\gamma$ of the
resulting domain graph, as the intersection of the edge sets $E(G_k)$
is not empty. 

One may wonder at this point why we ignore the edge weights when
computing the DFs. The reason is simply that experimentally, we observed in our examples that it leads to a decrease in accuracy
for a higher computing cost. Thus, it seems sufficient to consider
only the topological properties of the domain graph. 

Computing the DFs amounts to computing Pagerank
vectors and many different efficient algorithms
\cite{Berkhin05asurvey} have been developed after the Pagerank algorithm was first described \cite{Brin98}. One of the fastest algorithm currently known uses a Monte Carlo-based incremental approach and has a complexity of $\mathcal{O}(\frac{n \ln m}{\epsilon^2})$, where $n$ is the number of nodes, $m$ the number of selected edges and $\epsilon$ the desired precision \cite{bahmani2010}. We also observe that it may not be necessary to  reach full convergence, as we experimentally get a quasi-maximal accuracy when diffusing for a limited number of steps approximately equal to the directed diameter of the domain graph (\eg, 6 in the case of the domain graphs we used for our text classification experiments). 

Once the Pagerank vector of the domain graph has been computed, the
heuristic we use for reducing the dimension of the fingerprint vectors
is very fast, since it only consists in selecting a subset of the
vector indices. We note that there is a very high correlation between
the Pagerank vector of the domain graph and the component-wise average
of the fingerprint vectors (\eg, Spearman correlation of $0.92$ for
$\gamma = \e{-2}$ in the graph used for gender detection). We can thus get a good approximate of the Pagerank vector by computing the average of the fingerprint vectors. Incidentally, we also find a nearly perfect correlation between the component-wise average and component-wise variance of the fingerprint vectors in the studied graphs. This suggests that it may not be necessary after all to compute the Pagerank vector of the domain graph, in order to quickly identify the most central nodes. 

\section{Conclusion}

We have presented a method to generate fingerprint vectors for
data exhibiting associative properties. It is based on diffusion
processes over a domain graph and shows how to make dimensionality
reduction efficient and robust. The numerical vectors that get
generated can subsequently be used for classification or clustering. 

We applied our method to two classical text classification
problems with the same set of blogs. We showed that in both the case
of gender detection and of authorship attribution, \textit{Diffusion
  Fingerprints} provide a better accuracy and a greater resilience to
dimension reduction than equivalent bag-of-words vectors. 

The adaption and application of our method to the problem of metabolic 
subgraph extraction led to results that favorably compare with the state 
of the art in the field, establishing the power and flexibility of this 
elegant and conceptually simple method. The hybridization of this 
algorithm could in principle improve the quality of the results obtained above.

In the context of classification, we believe that DFs may prove useful not just for authorship detection but in many other domains, provided that a domain graph has been constructed. In the case of Free Association datasets for example \cite{fa98,dubuisson2013}, DF could be used to detect cultural shifts or psychological disorders of certain individuals. When studying social networks (\eg, friendship networks, co-author networks, online social networks, \dots), DF could be used to compare the social profiles of different members of a group. DF could also find fruitful application to Word Sense Disambiguation by enabling to generate distinctive contextual fingerprints for the words to be disambiguated. 


\bibliographystyle{acml2015}
\bibliography{biblio}

\begin{thebibliography}{38}
\providecommand{\natexlab}[1]{#1}
\providecommand{\url}[1]{\texttt{#1}}
\expandafter\ifx\csname urlstyle\endcsname\relax
  \providecommand{\doi}[1]{doi: #1}\else
  \providecommand{\doi}{doi: \begingroup \urlstyle{rm}\Url}\fi

\bibitem[Andersen et~al.(2006{\natexlab{a}})Andersen, Chung, and
  Lang]{Andersen:2006:LGP:1170136.1170528}
R.~Andersen, F.~Chung, and K.~Lang.
\newblock Local graph partitioning using pagerank vectors.
\newblock In \emph{Proceedings of the 47th Annual IEEE Symposium on Foundations
  of Computer Science}, FOCS '06, pages 475--486, Washington, DC, USA,
  2006{\natexlab{a}}. IEEE Computer Society.
\newblock ISBN 0-7695-2720-5.
\newblock \doi{10.1109/FOCS.2006.44}.

\bibitem[Andersen et~al.(2006{\natexlab{b}})Andersen, Chung, and
  Lang]{andersen2006}
R.~Andersen, F.~Chung, and K.~Lang.
\newblock Local graph partitioning using pagerank vectors.
\newblock In \emph{Foundations of Computer Science, 2006. FOCS'06. 47th Annual
  IEEE Symposium on}, pages 475--486. IEEE, 2006{\natexlab{b}}.

\bibitem[Antonov et~al.(2009)Antonov, Dietmann, Wong, and Mewes]{Antonov09}
A.~V. Antonov, S.~Dietmann, P.~Wong, and H.~W. Mewes.
\newblock Ticl – a web tool for network-based interpretation of compound
  lists inferred by high-throughput metabolomics.
\newblock \emph{FEBS Journal}, 276\penalty0 (7):\penalty0 2084--2094, 2009.
\newblock ISSN 1742-4658.
\newblock \doi{10.1111/j.1742-4658.2009.06943.x}.

\bibitem[Bahmani et~al.(2010)Bahmani, Chowdhury, and Goel]{bahmani2010}
B.~Bahmani, A.~Chowdhury, and A.~Goel.
\newblock Fast incremental and personalized pagerank.
\newblock \emph{Proceedings of the VLDB Endowment}, 4\penalty0 (3):\penalty0
  173--184, 2010.

\bibitem[Belkin and Niyogi(2001)]{belkin2001}
M.~Belkin and P.~Niyogi.
\newblock Laplacian eigenmaps and spectral techniques for embedding and
  clustering.
\newblock In \emph{NIPS}, volume~14, pages 585--591, 2001.

\bibitem[Belkin and Niyogi(2003)]{Belkin03laplacianeigenmaps}
M.~Belkin and P.~Niyogi.
\newblock Laplacian eigenmaps for dimensionality reduction and data
  representation.
\newblock \emph{Neural Computation}, 15:\penalty0 1373--1396, 2003.

\bibitem[Berkhin(2005)]{Berkhin05asurvey}
P.~Berkhin.
\newblock A survey on pagerank computing.
\newblock \emph{Internet Mathematics}, 2:\penalty0 73--120, 2005.

\bibitem[Breiman(2001)]{RF}
L.~Breiman.
\newblock Random forests.
\newblock \emph{Machine Learning}, 45\penalty0 (1):\penalty0 5--32, 2001.
\newblock ISSN 0885-6125.
\newblock \doi{10.1023/A:1010933404324}.

\bibitem[Brin and Page(1998)]{Brin98}
S.~Brin and L.~Page.
\newblock The anatomy of a large-scale hypertextual {Web} search engine.
\newblock \emph{Computer Networks and ISDN Systems}, 30\penalty0
  (1--7):\penalty0 107--117, 1998.

\bibitem[Bryll et~al.(2003)Bryll, Gutierrez-Osuna, and Quek]{bryll2003}
R.~Bryll, R.~Gutierrez-Osuna, and F.~Quek.
\newblock Attribute bagging: improving accuracy of classifier ensembles by
  using random feature subsets.
\newblock \emph{Pattern recognition}, 36\penalty0 (6):\penalty0 1291--1302,
  2003.

\bibitem[Coifman et~al.(2005)Coifman, Lafon, Lee, Maggioni, Nadler, Warner, and
  Zucker]{coifman2005}
R.~R. Coifman, S.~Lafon, A.~B. Lee, M.~Maggioni, B.~Nadler, F.~Warner, and
  S.~W. Zucker.
\newblock Geometric diffusions as a tool for harmonic analysis and structure
  definition of data: Diffusion maps.
\newblock \emph{Proceedings of the National Academy of Sciences of the United
  States of America}, 102\penalty0 (21):\penalty0 7426--7431, 2005.

\bibitem[Dubuisson(2014)]{df14}
J.~Dubuisson.
\newblock {D}iffusion {F}ingerprints -- {A}pplication demo to text
  classification, 2014.

\bibitem[Dubuisson et~al.(2013)Dubuisson, Eckmann, Scheible, and
  Sch\"{u}tze]{dubuisson2013}
J.~Dubuisson, J.-P. Eckmann, C.~Scheible, and H.~Sch\"{u}tze.
\newblock The topology of semantic knowledge.
\newblock In \emph{Proceedings of the 2013 Conference on Empirical Methods in
  Natural Language Processing}, pages 669--680, Seattle, Washington, USA,
  October 2013. Association for Computational Linguistics.

\bibitem[Dupont et~al.(2006)Dupont, Callut, Dooms, Monette, Deville,
  et~al.]{Dupont06}
P.~Dupont, J.~Callut, G.~Dooms, J.-N. Monette, Y.~Deville, et~al.
\newblock Relevant subgraph extraction from random walks in a graph.
\newblock \emph{Universit{\'e} catholique de Louvain, UCL/INGI, Number RR}, 7,
  2006.

\bibitem[Faust and van Helden(2012)]{Faust11}
K.~Faust and J.~van Helden.
\newblock Predicting metabolic pathways by sub-network extraction.
\newblock In Jacques van Helden, Ariane Toussaint, and Denis Thieffry, editors,
  \emph{Bacterial Molecular Networks}, volume 804 of \emph{Methods in Molecular
  Biology}, pages 107--130. Springer New York, 2012.
\newblock ISBN 978-1-61779-360-8.
\newblock \doi{10.1007/978-1-61779-361-5_7}.

\bibitem[Faust et~al.(2010)Faust, Dupont, Callut, and van Helden]{Faust10}
K.~Faust, P.~Dupont, J.~Callut, and J.~van Helden.
\newblock Pathway discovery in metabolic networks by subgraph extraction.
\newblock \emph{Bioinformatics}, 26\penalty0 (9):\penalty0 1211--1218, 2010.
\newblock \doi{10.1093/bioinformatics/btq105}.

\bibitem[Fouss et~al.(2007)Fouss, Pirotte, Renders, and Saerens]{fouss2007}
F.~Fouss, A.~Pirotte, J.-M. Renders, and M.~Saerens.
\newblock Random-walk computation of similarities between nodes of a graph with
  application to collaborative recommendation.
\newblock \emph{Knowledge and data engineering, ieee transactions on},
  19\penalty0 (3):\penalty0 355--369, 2007.

\bibitem[Freund and Schapire(1995)]{Freund:1995:DGO:646943.712093}
Y.~Freund and R.~Schapire.
\newblock A decision-theoretic generalization of on-line learning and an
  application to boosting.
\newblock In \emph{Proceedings of the Second European Conference on
  Computational Learning Theory}, EuroCOLT '95, pages 23--37, London, UK, UK,
  1995. Springer-Verlag.
\newblock ISBN 3-540-59119-2.

\bibitem[Gori et~al.(2007)Gori, Pucci, Roma, et~al.]{gori2007}
M.~Gori, A.~Pucci, V.~Roma, et~al.
\newblock Itemrank: A random-walk based scoring algorithm for recommender
  engines.
\newblock In \emph{IJCAI}, volume~7, pages 2766--2771, 2007.

\bibitem[Hotelling(1933)]{Hotel}
H.~Hotelling.
\newblock Analysis of a complex of statistical variables into principal
  components.
\newblock \emph{Journal of Educational Psychology}, 24:\penalty0 417--441,
  1933.

\bibitem[Kondor and Lafferty(2002)]{kondor2002}
R.~I. Kondor and J.~Lafferty.
\newblock Diffusion kernels on graphs and other discrete input spaces.
\newblock In \emph{ICML}, volume~2, pages 315--322, 2002.

\bibitem[Koppel et~al.(2011)Koppel, Schler, and Argamon]{ko11}
M.~Koppel, J.~Schler, and S.~Argamon.
\newblock Authorship attribution in the wild.
\newblock \emph{Language Resources and Evaluation}, 45\penalty0 (1):\penalty0
  83--94, 2011.

\bibitem[Mukherjee and Liu(2010)]{mukherjee2010}
A.~Mukherjee and B.~Liu.
\newblock Improving gender classification of blog authors.
\newblock In \emph{Proceedings of the 2010 conference on Empirical Methods in
  natural Language Processing}, pages 207--217. Association for Computational
  Linguistics, 2010.

\bibitem[Nadler et~al.(2005)Nadler, Lafon, Coifman, and
  Kevrekidis]{Nadler05diffusionmaps}
B.~Nadler, S.~Lafon, R.~Coifman, and I.~Kevrekidis.
\newblock Diffusion maps, spectral clustering and eigenfunctions of
  fokker-planck operators.
\newblock In \emph{in Advances in Neural Information Processing Systems 18},
  pages 955--962. MIT Press, 2005.

\bibitem[Nelson et~al.(2004)Nelson, McEvoy, and Schreiber]{fa98}
D.~Nelson, C.~McEvoy, and T.~Schreiber.
\newblock The {U}niversity of {S}outh {F}lorida free association, rhyme, and
  word fragment norms.
\newblock \emph{Behavior Research Methods, Instruments, \& Computers},
  36\penalty0 (3):\penalty0 402--407, 2004.

\bibitem[Page et~al.(1999)Page, Brin, Motwani, and Winograd]{page1999}
L.~Page, S.~Brin, R.~Motwani, and T.~Winograd.
\newblock The pagerank citation ranking: Bringing order to the web.
\newblock 1999.

\bibitem[Pearson(1901)]{PCA}
K.~Pearson.
\newblock On lines and planes of closest fit to systems of points in space.
\newblock \emph{Philosophical Magazine}, 2:\penalty0 559--572, 1901.

\bibitem[Perozzi et~al.(2014)Perozzi, Al-Rfou, and Skiena]{perozzi2014}
B.~Perozzi, R.~Al-Rfou, and S.~Skiena.
\newblock Deepwalk: Online learning of social representations.
\newblock In \emph{Proceedings of the 20th ACM SIGKDD international conference
  on Knowledge discovery and data mining}, pages 701--710. ACM, 2014.

\bibitem[Pons and Latapy(2006)]{pons2006}
P.~Pons and M.~Latapy.
\newblock Computing communities in large networks using random walks.
\newblock \emph{J. Graph Algorithms Appl.}, 10\penalty0 (2):\penalty0 191--218,
  2006.

\bibitem[Rajagopalan and Agarwal(2005)]{Rajagopalan05}
D.~Rajagopalan and P.~Agarwal.
\newblock Inferring pathways from gene lists using a literature-derived network
  of biological relationships.
\newblock \emph{Bioinformatics}, 21\penalty0 (6):\penalty0 788--793, 2005.
\newblock \doi{10.1093/bioinformatics/bti069}.

\bibitem[Roweis and Saul(2000)]{Roweis00nonlineardimensionality}
S.~Roweis and L.~Saul.
\newblock Nonlinear dimensionality reduction by locally linear embedding.
\newblock \emph{Science}, 290:\penalty0 2323--2326, 2000.

\bibitem[Schler et~al.(2006)Schler, Koppel, Argamon, and
  Pennebaker]{schler2006}
J.~Schler, M.~Koppel, S.~Argamon, and J.~W. Pennebaker.
\newblock Effects of age and gender on blogging.
\newblock In \emph{AAAI Spring Symposium: Computational Approaches to Analyzing
  Weblogs}, volume~6, pages 199--205, 2006.

\bibitem[Sch{\"o}lkopf and Smola(2002)]{scholkopf2002}
B.~Sch{\"o}lkopf and A.~J. Smola.
\newblock \emph{Learning with kernels: Support vector machines, regularization,
  optimization, and beyond}.
\newblock MIT press, 2002.

\bibitem[Scott et~al.(2005)Scott, Perkins, Bunnell, Pepin, Thomas, and
  Hallett]{Scott05}
M.~S. Scott, T.~Perkins, S.~Bunnell, F.~Pepin, D.~Y. Thomas, and M.~Hallett.
\newblock Identifying regulatory subnetworks for a set of genes.
\newblock \emph{Molecular \& Cellular Proteomics}, 4\penalty0 (5):\penalty0
  683--692, 2005.

\bibitem[Seroussi et~al.(2012)Seroussi, Bohnert, and
  Zukerman]{Seroussi:2012:AAA:2390665.2390728}
Y.~Seroussi, F.~Bohnert, and I.~Zukerman.
\newblock Authorship attribution with author-aware topic models.
\newblock In \emph{Proceedings of the 50th Annual Meeting of the Association
  for Computational Linguistics: Short Papers - Volume 2}, ACL '12, pages
  264--269, Stroudsburg, PA, USA, 2012. Association for Computational
  Linguistics.

\bibitem[Shi and Malik(2000)]{shi2000}
J.~Shi and J.~Malik.
\newblock Normalized cuts and image segmentation.
\newblock \emph{Pattern Analysis and Machine Intelligence, IEEE Transactions
  on}, 22\penalty0 (8):\penalty0 888--905, 2000.

\bibitem[Tenenbaum et~al.(2000)Tenenbaum, Silva, and Langford]{Tenenbaum2000}
J.~Tenenbaum, V.~Silva, and J.~Langford.
\newblock A global geometric framework for nonlinear dimensionality reduction.
\newblock \emph{Science}, 290:\penalty0 2319{--}2323, 2000.

\bibitem[Zien et~al.(2000)Zien, K{\"u}ffner, Zimmer, and Lengauer]{Zien00}
A.~Zien, R.~K{\"u}ffner, R.~Zimmer, and T.~Lengauer.
\newblock Analysis of gene expression data with pathway scores.
\newblock In \emph{Ismb}, volume~8, pages 407--417, 2000.

\end{thebibliography}

\end{document}